\def\year{2018}\relax
\documentclass[letterpaper]{article} 
\usepackage{aaai18}  
\usepackage{times}  
\usepackage{helvet}  
\usepackage{courier}  
\usepackage{url}  
\usepackage{graphicx}  

\usepackage{macros}

\begin{document}
%
\title{Efficient-UCBV: An Almost Optimal Algorithm using Variance Estimates}
\author{Subhojyoti Mukherjee${}^1$, K. P. Naveen${}^2$, Nandan
Sudarsanam${}^{3,4}$, and Balaraman Ravindran${}^{1,4}$\\
${}^{1}$Department of Computer Science \& Engineering, Indian Institute of
Technology Madras\\ 
${}^2$Department of Electrical Engineering, Indian Institute of
Technology Tirupati\\
${}^3$Department of Management Studies, Indian Institute of
Technology Madras\\
${}^4$ Robert Bosch Centre for Data Science and AI (RBC-DSAI), Indian Institute of
Technology Madras\\
${}^1$subho@cse.iitm.ac.in, ${}^2$naveenkp@iittp.ac.in, ${}^3$nandan@iitm.ac.in, ${}^4$ravi@cse.iitm.ac.in}

\maketitle

\begin{abstract}
We propose a novel variant of the UCB algorithm (referred to as Efficient-UCB-Variance (EUCBV)) for minimizing cumulative regret in the stochastic multi-armed bandit (MAB) setting. EUCBV incorporates the arm elimination strategy proposed in UCB-Improved \citep{auer2010ucb}, while taking into account the variance estimates to compute the arms' confidence bounds, similar to UCBV \citep{audibert2009exploration}. Through a theoretical analysis we establish that EUCBV incurs a \emph{gap-dependent} regret bound of {\scriptsize $O\left( \dfrac{K\sigma^2_{\max} \log (T\Delta^2 /K)}{\Delta}\right)$} after $T$ trials, where $\Delta$ is the minimal gap between optimal and sub-optimal arms; the above bound is an improvement over that of existing state-of-the-art UCB algorithms (such as UCB1, UCB-Improved, UCBV,  MOSS). Further, EUCBV incurs a \emph{gap-independent} regret bound of {\scriptsize $O\left(\sqrt{KT}\right)$}  which is an improvement over that of UCB1, UCBV and UCB-Improved, while being comparable with that of MOSS and OCUCB. Through an extensive numerical study we show that EUCBV significantly outperforms the popular UCB variants (like MOSS, OCUCB, etc.) as well as Thompson sampling and Bayes-UCB algorithms. 

\end{abstract}

\section{Introduction}
\label{sec:intro}
In this paper, we deal with the stochastic multi-armed bandit (MAB) setting. In its classical form, stochastic MABs represent a sequential learning problem where a learner is exposed to a finite set of actions (or arms) and needs to choose one of the actions at each timestep. After choosing (or pulling) an arm the learner  receives a reward, which is conceptualized as an independent random draw from stationary distribution associated with the selected arm. 
The mean of the reward distribution associated with an arm $i$ is denoted by $r_i$ whereas the mean of the reward distribution of the optimal arm $*$ is denoted by $r^*$ such that $r_i < r^*, \forall i\in \A$, where $\A$ is the set of arms such that $|\A|=K$. With this formulation the learner faces the task of balancing exploitation and exploration. In other words, should the learner pull the arm which currently has the best known estimates or explore arms more thoroughly to ensure that a correct decision is being made. The objective in the stochastic bandit problem is to minimize the cumulative regret, which is defined as follows:
\begin{align*}
R_{T}=r^{*}T - \sum_{i =1}^{K} r_{i}z_{i}(T),
\end{align*}
where $T$ is the number of timesteps, and  $z_{i}(T)$ is the number of times the algorithm has chosen arm $i$ up to timestep $T$.
The expected regret of an algorithm after $T$ timesteps can be written as,
\begin{align*}
\E[R_{T}]= \sum_{i=1}^{K} \E[z_i (T)] \Delta_i,
\end{align*}
where $\Delta_{i}=r^{*}-r_{i}$ is the gap between the means of the optimal arm and the $i$-th arm.


	In recent years the MAB setting has garnered extensive popularity because of its simple learning  model and its practical applications in a wide-range of industries, including, but not limited to, mobile channel allocations, online advertising and computer simulation games. 
	

\subsection{Related Work}
\label{sec:related}

	Bandit problems have been extensively studied in several earlier works such as \citet{thompson1933likelihood}, \citet{robbins1952some} and \citet{lai1985asymptotically}. \citet{lai1985asymptotically} established an asymptotic lower bound for the cumulative regret. Over the years stochastic MABs have seen several algorithms with strong regret guarantees. For further reference an interested reader can look into \citet{bubeck2012regret}. The upper confidence bound algorithms balance the exploration-exploitation dilemma by linking the uncertainty in estimate of an arm with the number of times an arm is pulled, and therefore ensuring sufficient exploration. One of the earliest among these algorithms is UCB1 \citep{auer2002finite}, which has a gap-dependent regret upper bound of  $O\left(\frac{K\log T}{\Delta}\right)$, where $\Delta = \min_{i:\Delta_i>0} \Delta_i$. This result is asymptotically order-optimal for the class of distributions considered. But, the worst case gap-independent regret bound of UCB1 is found to be  $O \left(\sqrt{KT\log T}\right)$. In the later work of \citet{audibert2009minimax}, the authors propose the MOSS algorithm and showed that the worst case gap-independent regret bound of MOSS is $O\left( \sqrt{KT} \right)$ which improves upon UCB1 by a factor of order $\sqrt{\log T}$. However, the gap-dependent regret of MOSS is $O\left( \frac{K^{2}\log\left(T\Delta^{2}/K\right)}{\Delta}\right)$ and in certain regimes, this can be worse than even UCB1 (see \citet{audibert2009minimax,lattimore2015optimally}).
	
	 The UCB-Improved algorithm, proposed in \citet{auer2010ucb}, is a round-based\footnote{An algorithm is \textit{round-based} if it pulls all the arms equal number of times in each round and then eliminates one or more arms that it deems  to be sub-optimal.} variant of UCB1, that 
incurs a gap-dependent regret bound of $O\left(\frac{K\log (T\Delta^{2})}{\Delta}\right)$, which is better than that of UCB1. On the other hand, the worst case gap-independent regret bound of UCB-Improved is $O\left(\sqrt{KT\log K}\right)$. Recently in \citet{lattimore2015optimally}, the authors showed that  the algorithm OCUCB achieves order-optimal gap-dependent regret bound of $O\left(\sum_{i=2}^{K}\frac{\log\left(T/H_i\right)}{\Delta_i}\right)$ where $H_i=\sum_{j=1}^{K}\min\left\lbrace \frac{1}{\Delta_i^2},\frac{1}{\Delta_j^2}\right\rbrace$, and a gap-independent regret bound of $O\left( \sqrt{KT}\right)$. This is the best known gap-dependent and gap-independent regret bounds in the stochastic MAB framework. However, unlike our proposed EUCBV algorithm, OCUCB does not take into account the variance of the arms; as a result, empirically  we find  that our algorithm outperforms OCUCB in all the environments considered. 

	In contrast to the above work, the UCBV \citep{audibert2009exploration} algorithm utilizes variance estimates to compute the confidence intervals for each arm. UCBV has a gap-dependent regret bound of $O\left(\frac{K\sigma_{\max}^{2}\log T}{\Delta}\right)$, where $\sigma_{\max}^{2}$ denotes the maximum variance among all the arms $i\in \A$. Its gap-independent regret bound can be inferred to be same as that of UCB1 i.e $O \left(\sqrt{KT\log T}\right)$. Empirically, \citet{audibert2009exploration} showed that UCBV outperforms UCB1 in several scenarios. 
	
	Another notable design principle which has recently gained a lot of popularity is the Thompson Sampling (TS) algorithm (\citep{thompson1933likelihood}, \citep{agrawal2011analysis})  and  Bayes-UCB (BU) algorithm \citep{kaufmann2012bayesian}. 
The TS algorithm maintains a posterior reward distribution for each arm; at each round, the algorithm samples values from these distributions and the arm corresponding to the highest sample value is chosen. Although TS is found to perform extremely well when the reward distributions are Bernoulli, it is established that with Gaussian priors the worst case regret can be as bad as $\Omega \left( \sqrt{KT\log T}\right)$ \citep{lattimore2015optimally}. The BU algorithm is an extension of the TS algorithm that takes quartile deviations into consideration while choosing arms.
	
	The final design principle we state is the information theoretic approach of DMED  \citep{honda2010asymptotically} and KLUCB \citep{garivier2011kl} algorithms. The algorithm KLUCB uses Kullbeck-Leibler divergence to compute the upper confidence bound for the arms. KLUCB is stable for a short horizon and is known to reach the \citet{lai1985asymptotically} lower bound in the special case of Bernoulli distribution. However, \citet{garivier2011kl} showed that KLUCB, MOSS and UCB1 algorithms are  empirically outperformed by UCBV in the exponential distribution as they do not take the variance of the arms into consideration.

\subsection{Our Contributions}
\label{sec:contri}
In this paper we propose the Efficient-UCB-Variance (henceforth referred to as EUCBV) algorithm for the stochastic MAB setting. EUCBV combines the approaches of UCB-Improved, CCB \citep{liu2016modification} and UCBV algorithms. EUCBV, by virtue of taking into account the empirical variance of the arms, exploration parameters  and non-uniform arm selection (as opposed to UCB-Improved), performs significantly better than the existing algorithms in the stochastic MAB setting. EUCBV outperforms UCBV \citep{audibert2009exploration} which also takes into account empirical variance but is less powerful than EUCBV because of the usage of exploration regulatory factor by EUCBV. Also, we carefully design the confidence interval term with the variance estimates along with the pulls allocated to each arm to balance the risk of eliminating the optimal arm against excessive optimism. Theoretically we refine the analysis of \citet{auer2010ucb} and prove that for $T\geq K^{2.4}$ our algorithm is order optimal and achieves a worst case gap-independent regret bound of $O\left( \sqrt{KT} \right)$ which is same as that of MOSS and OCUCB but better than that of UCBV, UCB1 and UCB-Improved. Also, the gap-dependent regret bound of EUCBV is better than UCB1, UCB-Improved and MOSS but is poorer than OCUCB. However, EUCBV's gap-dependent bound matches OCUCB in the worst case scenario when all the gaps are equal. Through our theoretical analysis we establish the exact values of the exploration parameters for the best performance of EUCBV. Our proof technique is highly generic and can be easily extended to other MAB settings. In Table \ref{tab:comp-bds} we show the regret bounds of different algorithms.

\begin{table}[!th]
\caption{Regret upper bound of different algorithms}
\label{tab:comp-bds}
\begin{center}
\begin{tabular}{p{3em}p{9em}p{7em}}
\toprule
Algorithm  &   \hspace*{1mm}Gap-Dependent & Gap-Independent \\
\hline
EUCBV		& $O\left( \dfrac{K\sigma_{\max}^{2}\log (\frac{T\Delta^2}{K})}{\Delta}\right)$ & $O\left(\sqrt{KT}\right)$\\
UCB1        & $O\left( \dfrac{K\log T}{\Delta} \right)$ & $O\left(\sqrt{KT\log T}\right)$ \\
UCBV        & $O\left( \dfrac{K\sigma_{\max}^{2}\log T}{\Delta} \right)$ & $O\left(\sqrt{KT\log T}\right)$ \\
UCB-Imp 		& $O\left( \dfrac{K\log (T\Delta^2)}{\Delta} \right)$ & $O\left(\sqrt{KT\log K}\right)$ \\
MOSS	     	& $O\left( \dfrac{K^2\log (T\Delta^2 /K)}{\Delta}\right)$ & $O\left(\sqrt{KT}\right)$\\
OCUCB     	& $O\left( \dfrac{K\log (T/ H_{i})}{\Delta}\right)$ & $O\left(\sqrt{KT}\right)$\\\midrule
\end{tabular}
\end{center}
\vspace*{-2em}
\end{table}

Empirically, we show that EUCBV, owing to its estimating the variance of the arms, exploration parameters and non-uniform arm pull, performs significantly better than MOSS, OCUCB, UCB-Improved, UCB1, UCBV, TS, BU, DMED, KLUCB and Median Elimination algorithms. Note that except UCBV, TS, KLUCB and BU (the last three with Gaussian priors) all the aforementioned algorithms do not take into account the empirical variance estimates of the arms. Also, for the optimal performance of TS, KLUCB and BU one has to have the prior knowledge of the type of distribution, but EUCBV requires no such prior knowledge. EUCBV is the first arm-elimination algorithm that takes into account the variance estimates of the arm for minimizing cumulative regret and thereby answers an open question raised by \citet{auer2010ucb}, where the authors conjectured that an UCB-Improved like arm-elimination algorithm can greatly benefit by taking into consideration the variance of the arms.  A similar variance based arm-elimination algorithm has been proposed before for minimizing the  expected loss in pure-exploration thresholding bandit setup in \citet{mukherjee2016}. Also, EUCBV is the first algorithm that follows the same proof technique of UCB-Improved and achieves a gap-independent regret bound of $O\left( \sqrt{KT} \right)$ thereby, closing the gap of UCB-Improved which achieved a gap-independent regret bound of $O\left( \sqrt{KT\log K} \right)$. 
	
	The rest of the paper is organized as follows. In section~\ref{sec:eucbv} we present the  EUCBV algorithm. Our main theoretical results are stated in section~\ref{sec:results}, while the proofs are established in   section \ref{sec:proofTheorem}. Section~\ref{sec:expt} contains results and discussions from our numerical experiments. We draw our conclusions in section \ref{sec:conc} and section \ref{sec:app} is Appendix (supplementary material).
	
	

\section{Algorithm: Efficient UCB Variance}
\label{sec:eucbv}
\algblock{ArmElim}{EndArmElim}
\algnewcommand\algorithmicArmElim{\textbf{\em Arm Elimination}}
 \algnewcommand\algorithmicendArmElim{}
\algrenewtext{ArmElim}[1]{\algorithmicArmElim\ #1}
\algrenewtext{EndArmElim}{\algorithmicendArmElim}

\algblock{ResParam}{EndResParam}
\algnewcommand\algorithmicResParam{\textbf{\em Reset Parameters}}
 \algnewcommand\algorithmicendResParam{}
\algrenewtext{ResParam}[1]{\algorithmicResParam\ #1}
\algrenewtext{EndResParam}{\algorithmicendResParam}

\begin{algorithm}[!th]
\caption{EUCBV}
\label{alg:eucbv}
\begin{algorithmic}
\State {\bf Input:} Time horizon $T$, exploration parameters $\rho$ and $\psi$.
\State {\bf Initialization:} Set $m:=0$, $B_{0}:=\mathcal{A}$, $\epsilon_{0}:=1$, $M=\big \lfloor \frac{1}{2}\log_{2} \frac{T}{e}\big\rfloor$, $n_{0}=\big\lceil\frac{\log{(\psi T\epsilon_{0}^{2})}}{2\epsilon_{0}}\big\rceil$ and  $N_{0}=Kn_{0}$.
\State Pull each arm once
\For{$t=K+1,..,T$}	
\State Pull arm $i\in \argmax_{j\in B_{m}}\bigg\lbrace \hat{r}_{j} + \sqrt{\frac{\rho(\hat{v}_{j}+2)\log{(\psi T\epsilon_{m})}}{4 z_{j}}} \bigg\rbrace$, where $z_j$ is the number of times arm $j$ has been pulled.
\ArmElim
\State For each arm $i \in B_{m}$, remove arm $i$ from $B_{m}$ if,
\begin{align*}
 \hat{r}_{i} + & \sqrt{\frac{\rho(\hat{v}_{i}+2)\log{(\psi T\epsilon_{m})}}{4 z_{i}}}  \\ 
 & < \max_{{j}\in B_{m}}\bigg\lbrace\hat{r}_{j} -\sqrt{\frac{\rho(\hat{v}_{j}+2)\log{(\psi T\epsilon_{m})}}{4 z_{j}}} \bigg\rbrace
\end{align*}
\EndArmElim

\If{$t\geq N_{m}$ and $m\leq M$}
\ResParam
\State $\epsilon_{m+1}:=\frac{\epsilon_{m}}{2}$\vspace{0.5ex}
\State $B_{m+1}:=B_{m}$
\State $n_{m+1}:=\bigg\lceil\frac{\log{(\psi T\epsilon_{m+1}^{2})}}{2\epsilon_{m+1}}\bigg\rceil$
\State $N_{m+1}:=t+|B_{m+1}| n_{m+1}$
\State $m:=m+1$
\EndResParam
\EndIf
\State Stop if $|B_{m}|=1$ and pull ${i}\in B_{m}$ till $T$ is reached.
\EndFor
\end{algorithmic}
\end{algorithm}
\textbf{2.1 Notations:} We denote the set of arms by $\A$, with the individual arms labeled $i$, where  $i=1,\ldots,K$. We denote an arbitrary round of EUCBV by $m$. For simplicity, we assume that the optimal arm is unique and denote it by ${*}$. We denote the sample mean of the rewards for an arm $i$ at time instant $t$ by $\hat{r}_{i}(t)=\frac{1}{z_{i}(t)}\sum_{\ell=1}^{z_i(t)} X_{i,\ell}$, where $X_{i,\ell}$ is the reward sample received when arm $i$ is pulled for the $\ell$-th time, and $z_i(t)$ is the number of times arm $i$ has been pulled until timestep $t$. We denote the true variance of an arm by $\sigma_i^{2}$ while $\hat{v}_{i}(t)$ is the estimated variance, i.e., $\hat{v}_{i}(t)=\frac{1}{z_i(t)}\sum_{\ell=1}^{z_{i}(t)}(X_{i,\ell}-\hat{r}_{i})^{2}$. Whenever there is no ambiguity about the underlaying  time index $t$, for simplicity we neglect $t$ from the notations and simply use  $\hat{r}_i, \hat{v}_i,$ and $z_i$ to denote the respective quantities. We assume the rewards of all arms are bounded in $[0,1]$.

\textbf{2.2 The algorithm:} Earlier round-based arm elimination algorithms like Median Elimination \citep{even2006action} and UCB-Improved mainly suffered from two basic problems: \\
\begin{inparaenum}[\bfseries(i)]
\item \textit{Initial exploration:} Both of these algorithms pull each arm equal number of times in each round, and hence waste a significant number of pulls in initial explorations. \\
\item \textit{Conservative arm-elimination:} In UCB-Improved, arms are eliminated conservatively, i.e, only after $\epsilon_{m}<\frac{\Delta_{i}}{2}$, 
where the quantity $\epsilon_{m}$ is initialized to $1$ and halved after every round. In the worst case scenario when $K$ is large, and the gaps are uniform  ($r_{1}=r_{2}=\cdots=r_{K-1}<r^{*}$) and small this results in very high regret.\\
\end{inparaenum}
\\
	The EUCBV algorithm, which is mainly based on the arm elimination technique of the UCB-Improved algorithm,  remedies these by employing exploration regulatory factor $\psi$ and arm elimination parameter $\rho$ for aggressive elimination of sub-optimal arms. Along with these, similar to CCB \citep{liu2016modification} algorithm, EUCBV uses optimistic greedy sampling whereby at every timestep it only pulls the arm with the highest upper confidence bound rather than pulling all the arms equal number of times in each round. Also, unlike the UCB-Improved, UCB1, MOSS and OCUCB algorithms (which are based on mean estimation) EUCBV employs mean and variance estimates (as in \citet{audibert2009exploration}) for arm elimination. Further, we allow for arm-elimination at every time-step, which is in contrast to the earlier work (e.g., \citet{auer2010ucb}; \citet{even2006action}) where the arm elimination takes place only at the end of the respective exploration rounds.

\section{Main Results} 
\label{sec:results}
The main result of the paper is presented in the following theorem, where we establish a regret upper bound for the proposed EUCBV  algorithm. 
\begin{theorem}[\textbf{\textit{Gap-Dependent Bound}}]
\label{Result:Theorem:1}
For $T\geq K^{2.4}$, $\rho=\frac{1}{2}$ and $\psi=\frac{T}{K^2}$, the regret $R_T$ for EUCBV satisfies
\begin{align*}
\E [R_{T}] \leq &\sum\limits_{i\in \A :\Delta_{i} > b}\bigg\lbrace \dfrac{C_0 K^{4}}{T^{\frac{1}{4}}} + \bigg(\Delta_{i}+\dfrac{320\sigma_i^2\log{(\frac{T\Delta_{i}^{2}}{K})}}{\Delta_{i}}\bigg)\bigg \rbrace\\ 
  & +\sum\limits_{i\in \A :0 < \Delta_{i}\leq b} \dfrac{C_2 K^{4}}{T^{\frac{1}{4}}} + \max_{i\in \A :0 < \Delta_{i}\leq b}\Delta_{i}T.
\end{align*}

for all $b\geq\sqrt{\frac{e}{T}}$ and $C_0, C_2$ are integer constants. 
\end{theorem}

\begin{proof}[Outline]
The proof is along the lines of the technique in \citet{auer2010ucb}. It comprises of three modules. In the first module we prove the necessary conditions for arm elimination within a specified number of rounds. However, here we require some additional technical results (see Lemma~\ref{proofTheorem:Lemma:1} and Lemma~\ref{proofTheorem:Lemma:2}) to bound the length of the confidence intervals. Further, note that our algorithm combines the variance-estimate based approach of \citet{audibert2009exploration} with the arm-elimination technique of \citet{auer2010ucb} (see Lemma~\ref{proofTheorem:Lemma:3}). Also, while \citet{auer2010ucb} uses Chernoff-Hoeffding bound to derive their regret bound whereas in our work we use  Bernstein inequality (as in \citet{audibert2009exploration}) to obtain the bound. To bound the probability of the non-uniform arm selection before it gets eliminated we use Lemma~\ref{proofTheorem:Lemma:4} and Lemma~\ref{proofTheorem:Lemma:5}. In the second module we bound the number of pulls required if an arm is eliminated on or before a particular number of rounds. Note that the number of pulls allocated in a round $m$ for each arm is $n_{m}:=\bigg\lceil\frac{\log{(\psi T\epsilon_{m}^{2})}}{2\epsilon_{m}}\bigg\rceil$ which is much lower than the number of pulls of each arm required by UCB-Improved or Median-Elimination. We introduce the variance term in the most significant term in the bound by Lemma~\ref{proofTheorem:Lemma:6}. Finally, the third module deals with case of bounding the regret, given that a sub-optimal arm eliminates the optimal arm.
\hfill $\blacksquare$
\end{proof}

\emph{Discussion:} From the above result we see that the most significant term in the gap-dependent bound is of the order $O\left(\frac{K\sigma^2_{\max}\log{(T\Delta^{2}/K)}}{\Delta}\right)$ which is better than the existing results for UCB1, UCBV, MOSS and UCB-Improved (see Table~\ref{tab:comp-bds}). Also, like UCBV, this term scales with the variance. \citet{audibert2010best} have defined the term $H_1=\sum_{i=1}^{K}\frac{1}{\Delta_i^2}$, which is referred to as the hardness of a problem; \citet{bubeck2012regret} have conjectured that the gap-dependent regret upper bound can match $O\left(\frac{K\log{(T/H_1)}}{\Delta}\right)$. However, in  \citet{lattimore2015optimally} it is proved that the gap-dependent regret bound cannot be lower than $O\left(\sum_{i=2}^{K}\frac{\log\left(T/H_i\right)}{\Delta_i}\right)$, where $H_i=\sum_{j=1}^{K}\min\left\lbrace \frac{1}{\Delta_i^2},\frac{1}{\Delta_j^2}\right\rbrace$ (OCUCB proposed in \citet{lattimore2015optimally} achieves this bound). Further, in \citet{lattimore2015optimally} it is shown that only in the worst case scenario when all the gaps are equal (so that $H_1=H_{i}=\sum_{i=1}^{K}\frac{1}{\Delta^2}$) the above two bounds match. In the latter scenario, considering $\sigma^2_{\max}\leq \frac{1}{4}$ as all rewards are bounded in $[0,1]$, we see that the gap-dependent bound of EUCBV simplifies to $O\left(\frac{K\log{(T/H_1)}}{\Delta}\right)$, thus matching the gap-dependent bound of OCUCB which is order optimal.

Next, we specialize the result of Theorem \ref{Result:Theorem:1} in Corollary \ref{Result:Corollary:1} to  obtain the gap-independent worst case regret bound. 


\begin{corollary}[\textbf{\textit{Gap-Independent Bound}}]
\label{Result:Corollary:1}
When the gaps of all the sub-optimal arms are identical, i.e., $\Delta_i =\Delta = \sqrt{\frac{K\log K}{T}}>\sqrt{\frac{e}{T}}, \forall i\in \A$ and $C_3$ being an integer constant, the
regret of EUCBV is upper bounded by the following gap-independent expression:
\begin{align*}
	\E[R_{T}]\leq  \dfrac{C_3 K^5}{T^{\frac{1}{4}}} + 80\sqrt{KT}.
\end{align*}	
\end{corollary}
	
The proof is given in Appendix \ref{App:Corollary:1}.

\emph{Discussion:} In the non-stochastic scenario, \citet{auer2002nonstochastic} showed that the bound on the cumulative regret for EXP-4 is $O\left(\sqrt{KT\log K}\right)$. However, in the stochastic case, UCB1 proposed in \citet{auer2002finite} incurred a regret of order of  $O\left(\sqrt{KT\log T}\right)$ which is clearly improvable. From the above result we see that in the gap-independent bound of EUCBV the most significant term is $O\left(\sqrt{KT}\right)$ which  matches the upper bound of MOSS and OCUCB, and is better than UCB-Improved, UCB1 and UCBV (see Table~\ref{tab:comp-bds}).


%
%

\section{Proofs}
\label{sec:proofTheorem}
We first present a few technical lemmas that are required  to prove the result in Theorem \ref{Result:Theorem:1}.

\begin{lemma}
\label{proofTheorem:Lemma:1}
If $T\geq K^{2.4}$, $\psi=\frac{T}{ K^2}$, $\rho=\frac{1}{2}$ and $m\leq \frac{1}{2} \log_2\left(\frac{T}{e}\right) $, then,
\begin{align*}
\dfrac{\rho m \log(2)}{\log(\psi T) - 2m\log( 2)} \leq \frac{3}{2}.
\end{align*}
\end{lemma}

\begin{lemma}
\label{proofTheorem:Lemma:2}
If $T\geq K^{2.4}$, $\psi=\frac{T}{ K^2}$, $\rho =\frac{1}{2}$, $m_i = min\lbrace m|\sqrt{4\epsilon_{m} } < \frac{\Delta_i}{4} \rbrace $ and $c_{i} =\sqrt{\frac{\rho (\hat{v}_i + 2)\log (\psi T\epsilon_{m_{i}})}{4 z_i}}$, then,
\center $c_{i} < \frac{\Delta_i}{4}$.
\end{lemma}

\begin{lemma}
\label{proofTheorem:Lemma:3}
If $m_i = min\lbrace m|\sqrt{4\epsilon_{m} } < \frac{\Delta_i}{4} \rbrace $,  $c_{i} = \sqrt{\frac{\rho (\hat{v}_i + 2) \log (\psi T\epsilon_{m_{i}})}{4 z_{i}}}$ and $n_{m_i} = \frac{\log{(\psi T\epsilon_{m_{i}})}}{2\epsilon_{m_{i}}}$ then we can show that in the $m_i$-th round,
\begin{align*}
\mathbb{P}(\hat{r}_{i}> r_{i} + c_{i})\le \dfrac{2}{(\psi  T\epsilon_{m_{i}})^{\frac{3\rho}{2}}}.
\end{align*}
\end{lemma}


\begin{lemma}
\label{proofTheorem:Lemma:4}
If $m_i = min\lbrace m|\sqrt{4\epsilon_{m} } < \frac{\Delta_i}{4} \rbrace $, $\psi=\frac{T}{ K^2}$, $\rho=\frac{1}{2}$, $c_{i} =\sqrt{\frac{\rho(\hat{v}_i + 2)\log (\psi T\epsilon_{m_{i}})}{4 z_{i}}}$ and $n_{m_i}=\frac{\log{(\psi T\epsilon_{m_{i}}^{2})}}{2\epsilon_{m_{i}}}$ then in the $m_i$-th round, 
\begin{align*}
\Pb\lbrace c^{*} > c_i \rbrace  \leq \dfrac{182 K^4}{T^{\frac{5}{4}}\sqrt{\epsilon_{m_i}}}.
\end{align*}
\end{lemma}

\begin{lemma}
\label{proofTheorem:Lemma:5}
If $m_i = min\lbrace m|\sqrt{4\epsilon_{m} } < \frac{\Delta_i}{4} \rbrace $,$\psi=\frac{T}{ K^2}$, $\rho=\frac{1}{2}$, $c_{i} =\sqrt{\frac{\rho (\hat{v}_i + 2)\log (\psi T\epsilon_{m_{i}})}{4 z_i}}$ and $n_{m_i}=\frac{\log{(\psi T\epsilon_{m_{i}}^{2})}}{2\epsilon_{m_{i}}}$ then in the $m_i$-th round, 
\begin{align*}
\Pb\lbrace z_i < n_{m_i} \rbrace  \leq \dfrac{182 K^4}{T^{\frac{5}{4}}\sqrt{\epsilon_{m_i}}}.
\end{align*}
\end{lemma}



\begin{lemma}
\label{proofTheorem:Lemma:6}
For two integer constants $c_1$ and $c_2$, if $20 c_1 \leq c_2$ then,
\begin{align*}
c_1 \frac{4\sigma_i^2 + 4}{\Delta_i}\log\bigg( \frac{T\Delta_i^2}{K}\bigg) \leq c_2 \frac{\sigma_i^2}{\Delta_i}\log\bigg( \frac{T\Delta_i^2}{K}\bigg).
\end{align*}
\end{lemma}

%
%

The proofs of lemmas \ref{proofTheorem:Lemma:1} - \ref{proofTheorem:Lemma:6} can be found in Appendix ~\ref{App:Lemma:1}, ~\ref{App:Lemma:2}, ~\ref{App:Lemma:3}, ~\ref{App:Lemma:4}, ~\ref{App:Lemma:5} and
 ~\ref{App:Lemma:6} respectively.


\subsection*{Proof of Theorem 1}
\label{sec:proofTheorem:Theorem1}
\begin{customproof}{1}
For each sub-optimal arm ${i}\in\mathcal{A}$, let $m_{i}=\min{\left\lbrace m|\sqrt{4\epsilon_{m_i}} < \frac{\Delta_{i}}{4}\right\rbrace}$. Also, let $\A^{'}=\lbrace i\in \A: \Delta_{i} > b \rbrace$ and $\A^{''}=\lbrace i\in \A: \Delta_{i} > 0 \rbrace$. Note that as all rewards are bounded in $[0,1]$, it implies that $0\leq \sigma_i^2 \leq \frac{1}{4},\forall i\in \A$. Now, as in \citet{auer2010ucb}, we bound the regret under the following two cases: 
\begin{itemize}
\item {Case $(a)$}: some sub-optimal arm ${i}$ is not eliminated in round $m_{i}$ or before and the optimal arm ${*}\in B_{m_{i}}$
\item {Case $(b)$}: an arm ${i}\in B_{m_i}$ is eliminated in round $m_{i}$ (or before), or there is no optimal arm $*\in B_{m_i}$
\end{itemize} 
The details of each case are contained in the following sub-sections.


\textbf{Case $(a)$:}
For simplicity, let $c_{i} := \sqrt{\frac{\rho (\hat{v}_i + 2) \log (\psi T\epsilon_{m_{i}})}{4 z_{i}}}$ denote the length of the confidence interval corresponding to arm $i$ in round $m_i$. Thus, in round $m_i$ (or before) whenever $z_i \geq n_{m_{i}}\ge\frac{\log{(\psi T\epsilon_{m_{i}}^{2})}}{2\epsilon_{m_{i}}}$, by applying Lemma \ref{proofTheorem:Lemma:2} we obtain $c_{i} < \frac{\Delta_{i}}{4}$.
Now, the sufficient conditions for arm $i$ to get eliminated by an optimal arm in round $m_i$ is given by
	\begin{eqnarray}
	\hat{r}_{i} \leq r_{i} + c_{i} \text{, } 
 	\hat{r}^{*} \geq r^{*} - c^{*} \text{, } c_{i} \geq c^* \text{ and } z_i \geq n_{m_i} \label{eq:armelim-casea}.
	\end{eqnarray}

Indeed, in round $m_i$ suppose (\ref{eq:armelim-casea}) holds, then we have
  \begin{align*}
\hat{r}_{i} + c_{i}&\leq r_{i} + 2c_{i} 
= r_{i} + 4c_{i} - 2c_{i} \\
 &< r_{i} + \Delta_{i} - 2c_{i}
 \leq r^{*} -2c^{*} 
 \leq \hat{r}^{*} - c^{*}
  \end{align*}
  so that a sub-optimal arm ${i} \in \A^{'}$ gets eliminated.	
Thus, the probability of the complementary event of these four conditions in (\ref{eq:armelim-casea}) yields a bound on the probability that arm $i$ is not eliminated in round $m_i$. Following the proof of Lemma 1 of \citet{audibert2009exploration} we can show that a bound on the complementary of the first condition is given by,

\begin{align}
\mathbb{P}(\hat{r}_{i}> r_{i} + c_{i})
&\leq \mathbb{P}\left( \hat{r}_{i} > r_{i}+ \bar{c}_i\right) 
+ \mathbb{P}\left( \hat{v}_{i}\geq \sigma_{i}^{2}+\sqrt{\epsilon_{m_{i}}}\right)\label{eq:prob_eq2}
\end{align}
where 
\begin{align*}
\bar{c}_i=\sqrt{\dfrac{\rho (\sigma_{i}^{2}+\sqrt{\epsilon_{m_{i}}} + 2)\log(\psi T\epsilon_{m_{i}})}{4n_{m_i}}}.
\end{align*}

From Lemma \ref{proofTheorem:Lemma:3} we can show that $\mathbb{P}(\hat{r}_{i}> r_{i} + c_{i})\leq\mathbb{P}\left( \hat{r}_{i} > r_{i}+ \bar{c}_i\right) + \mathbb{P}\left( \hat{v}_{i}\geq \sigma_{i}^{2}+\sqrt{\epsilon_{m_{i}}}\right) \leq \frac{2}{(\psi  T\epsilon_{m_{i}})^{\frac{3\rho}{2}}}$. Similarly, $\mathbb{P}\lbrace\hat{r}^{*} < r^{*} - c^{*}\rbrace \leq \frac{2}{(\psi  T\epsilon_{m_{i}})^{\frac{3\rho}{2}}}$. Summing the above two contributions, the probability that a sub-optimal arm ${i}$ is not eliminated on or before $m_{i}$-th round by the first two conditions in  (\ref{eq:armelim-casea}) is,  
\begin{eqnarray}
\bigg(\dfrac{4}{(\psi T\epsilon_{m_{i}})^{\frac{3\rho}{2}}} \bigg). \label{eq:arm:elim:c1}
\end{eqnarray}

Again, from Lemma \ref{proofTheorem:Lemma:4} and Lemma \ref{proofTheorem:Lemma:5} we can bound the probability of the  complementary of the event $c_{i} \geq c^* $ and $ z_i \geq n_{m_i}$ by,

\begin{eqnarray}
\dfrac{182 K^4}{T^{\frac{5}{4}}\sqrt{\epsilon_{m_i}}} + \dfrac{182 K^4}{T^{\frac{5}{4}}\sqrt{\epsilon_{m_i}}}\leq \dfrac{364 K^4}{T^{\frac{5}{4}}\sqrt{\epsilon_{m_i}}}. \label{eq:arm:elim:c2}
\end{eqnarray}

Also, for eq. $(\ref{eq:arm:elim:c1})$ we can show that for any $\epsilon_{m_i}\in[\sqrt{\frac{e}{T}},1]$
\begin{eqnarray}
\bigg(\dfrac{4}{(\psi T\epsilon_{m_{i}})^{\frac{3\rho}{2}}} \bigg) &\overset{(a)}{\leq} \bigg(\dfrac{4}{(\frac{T^2}{K^2}\epsilon_{m_{i}})^{\frac{3}{4}}} \bigg)\leq \bigg(\dfrac{4 K^{\frac{3}{2}}}{(T^\frac{3}{2} \epsilon_{m_i}^{\frac{1}{4}}\sqrt{\epsilon_{m_{i}}})}\bigg) \nonumber \\
&\overset{(b)}{\leq} \bigg(\dfrac{4 K^{\frac{3}{2}}}{(T^{\frac{3}{2}-\frac{1}{8}}\sqrt{\epsilon_{m_{i}}})}  \bigg)
\leq \dfrac{4 K^4}{T^{\frac{5}{4}}\sqrt{\epsilon_{m_i}}}. \label{eq:arm:elim:c3}
\end{eqnarray}

Here, in $(a)$ we substitute the values of $\psi$ and $\rho$ and $(b)$ follows from the identity $\epsilon_{m_i}^{\frac{1}{4}}\geq (\frac{e}{T})^{\frac{1}{8}} $ as $\epsilon_{m_i}\geq \sqrt{\frac{e}{T}}$.

Summing up over all arms in $\A^{'}$ and bounding the regret for all the \textit{four} arm elimination conditions in (\ref{eq:armelim-casea}) by $(\ref{eq:arm:elim:c2}) + (\ref{eq:arm:elim:c3})$ for each arm $i\in \A^{'}$ trivially by $T\Delta_{i}$, we obtain
	\begin{align*}
&\sum_{i\in \A^{'}}\bigg(\dfrac{4 K^4 T\Delta_i}{T^{\frac{5}{4}}\sqrt{\epsilon_{m_i}}}\bigg) + \sum_{i\in \A^{'}}\bigg(\dfrac{364 K^4 T\Delta_i}{T^{\frac{5}{4}}\sqrt{\epsilon_{m_i}}}\bigg)\\
&\overset{(a)}{\leq}\sum_{i\in \A^{'}}\bigg(\dfrac{368 K^4 T\Delta_{i}}{T^{\frac{5}{4}}\left(\frac{\Delta_{i}^{2}}{4.16}\right)^{\frac{1}{2}}}\bigg)
\overset{(b)}{\leq} \sum_{i\in \A^{'}}\bigg(\dfrac{C_1 K^4}{(T)^{\frac{1}{4}}}\bigg).\\  
	\end{align*}

Here, $(a)$ happens because $\sqrt{4\epsilon_{m_i}} < \frac{\Delta_i}{4}$, and in $(b)$, $C_1$ denotes a constant integer value.\\

\textbf{Case $(b)$:} Here, there are two sub-cases to be considered.

\noindent
\textbf{Case $(b1)$ (\textit{${*}\in B_{m_{i}}$ and each ${i}\in \A^{'}$ is  eliminated on or before $m_{i}$ }): } Since we are eliminating a sub-optimal arm ${i}$ on or before round $m_{i}$, it is pulled no longer than, 
 \begin{align*}
 z_{i} < \bigg\lceil\dfrac{\log{(\psi T\epsilon_{m_{i}}^{2})}}{2\epsilon_{m_{i}}}\bigg\rceil
 \end{align*}
So, the total contribution of ${i}$ until round $m_{i}$ is given by, 
\begin{align*}
&\Delta_{i}\bigg\lceil\dfrac{\log{(\psi T\epsilon_{m_{i}}^{2})}}{2\epsilon_{m_{i}}}\bigg\rceil
\overset{(a)}{\leq}    \Delta_{i}\bigg\lceil\dfrac{\log{(\psi T(\dfrac{\Delta_{i}}{16 \times 256})^{4})}}{2(\dfrac{\Delta_{i}}{4\sqrt{4}})^{2}}\bigg\rceil \\
&\leq   \Delta_{i}\bigg(1+\dfrac{32\log{(\psi T(\dfrac{\Delta_{i}^{4}}{16384})}}{\Delta_{i}^{2}}\bigg)
\leq \Delta_{i}\bigg(1+\dfrac{32\log{(\psi T\Delta_{i}^{4})}}{\Delta_{i}^{2}}\bigg) .
\end{align*} 

Here, $(a)$ happens because $\sqrt{4\epsilon_{m_{i}}} < \frac{\Delta_{i}}{4}$. Summing over all arms in $\A^{'}$ the total regret is given by, 
\begin{align*}
&\sum_{i\in \A^{'}}\Delta_{i}\bigg(1+\dfrac{32\log{(\psi T\Delta_{i}^{4}})}{\Delta_{i}^{2}}\bigg) = \sum_{i\in \A^{'}}\bigg(\Delta_{i} +\dfrac{32\log{(\psi T\Delta_{i}^{4}})}{\Delta_{i}}\bigg) \\
&\overset{(a)}{\leq} \sum_{i\in \A^{'}} \left(\Delta_{i}+\dfrac{64\log{( \frac{T\Delta_{i}^{2}}{K})}}{\Delta_{i}}\right)\\
&\overset{(b)}{\leq} \sum_{i\in \A^{'}} \left(\Delta_{i} +\dfrac{16(4\sigma_i^2 + 4)\log{( \frac{T\Delta_{i}^{2}}{K})}}{\Delta_{i}}\right)\\
&
\overset{(c)}{\leq} \sum_{i\in \A^{'}} \left(\Delta_{i} +\dfrac{320\sigma_i^2\log{( \frac{T\Delta_{i}^{2}}{K})}}{\Delta_{i}}\right).\\
\end{align*}

We obtain $(a)$ by substituting the value of $\psi$, $(b)$ from $0\leq\sigma_i^2 \leq\frac{1}{4},\forall i\in \A$ and $(c)$ from Lemma \ref{proofTheorem:Lemma:6}.\\

\noindent
\textbf{Case $(b2)$ (\textit{Optimal arm ${*}$ is eliminated by a sub-optimal arm):  }} Firstly, if conditions of Case $a$ holds then the optimal arm ${*}$ will not be eliminated in round $m=m_{*}$ or it will lead to the contradiction that $r_{i}>r^{*}$. In any round $m_{*}$, if the optimal arm ${*}$ gets eliminated then for any round from $1$ to $m_{j}$ all arms ${j}$ such that $m_{j}< m_{*}$ were eliminated according to assumption in Case $a$. Let the arms surviving till $m_{*}$ round be denoted by $\A^{'}$. This leaves any arm $a_{b}$ such that $m_{b}\geq m_{*}$ to still survive and eliminate arm ${*}$ in round $m_{*}$. Let such arms that survive ${*}$ belong to $\A^{''}$. Also maximal regret per step after eliminating ${*}$ is the maximal $\Delta_{j}$ among the remaining arms ${j}$ with $m_{j}\geq m_{*}$.  Let $m_{b}=\min\left\lbrace m|\sqrt{4\epsilon_{m}}<\frac{\Delta_{b}}{4}\right\rbrace$. Hence, the maximal regret after eliminating the arm ${*}$ is upper bounded by, 

\begin{align*}
&\sum_{m_{*}=0}^{max_{j\in \A^{'}}m_{j}}\sum_{i\in \A^{''}:m_{i}>m_{*}}\bigg(\dfrac{368 K^4}{(T^{\frac{5}{4}}\sqrt{\epsilon_{m_{*}}})} \bigg).T\max_{j\in \A^{''}:m_{j}\geq m_{*}}{\Delta}_{j}\\
&\leq\sum_{m_{*}=0}^{max_{j\in \A^{'}}m_{j}}\sum_{i\in \A^{''}:m_{i}>m_{*}}\bigg(\dfrac{368 K^4 \sqrt{4}}{(T^{\frac{5}{4}}\sqrt{\epsilon_{m_{*}}})} \bigg).T.4\sqrt{\epsilon_{m_{*}}}\\
&\overset{(a)}{\leq}\sum_{m_{*}=0}^{max_{j\in \A^{'}}m_{j}}\sum_{i\in \A^{''}:m_{i}>m_{*}}\bigg(\dfrac{C_2 K^4}{T^{\frac{1}{4}}\epsilon_{m_{*}}^{\frac{1}{2}-\frac{1}{2}}} \bigg)\\
&\leq\sum_{i\in \A^{''}:m_{i}>m_{*}}\sum_{m_{*}=0}^{\min{\lbrace m_{i},m_{b}\rbrace}}\bigg(\dfrac{C_2 K^4}{T^{\frac{1}{4}}} \bigg)\\
&\leq\sum_{i\in \A^{'}}\bigg(\dfrac{C_2 K^4}{T^{\frac{1}{4}}} \bigg)+\sum_{i\in \A^{''}\setminus \A^{'}}\bigg(\dfrac{C_2 K^4}{T^{\frac{1}{4}}} \bigg).\\
\end{align*}
Here at $(a)$, $C_2$ denotes an integer constant.

Finally, summing up the regrets in \textbf{Case a} and \textbf{Case b}, the total regret is given by
\begin{align*}
\E [R_{T}] \leq &\sum\limits_{i\in \A :\Delta_{i} > b}\bigg\lbrace \dfrac{C_0 K^{4}}{T^{\frac{1}{4}}} + \bigg(\Delta_{i}+\dfrac{320\sigma_i^2\log{(\frac{T\Delta_{i}^{2}}{K})}}{\Delta_{i}}\bigg)\bigg \rbrace\\ 
  & +\sum\limits_{i\in \A :0 < \Delta_{i}\leq b} \dfrac{C_2 K^{4}}{T^{\frac{1}{4}}} + \max_{i\in \A :0 < \Delta_{i}\leq b}\Delta_{i}T
\end{align*}

where $C_0, C_1, C_2$ are integer constants s.t. $C_0 = C_1 + C_2$.
\end{customproof}

\section{Experiments}
\label{sec:expt}
In this section, we conduct extensive empirical evaluations of EUCBV against several other popular MAB  algorithms. We use expected cumulative regret as the metric of comparison. The comparison is conducted against the following algorithms: KLUCB+ \citep{garivier2011kl}, DMED \citep{honda2010asymptotically}, MOSS \citep{audibert2009minimax}, UCB1 \citep{auer2002finite}, UCB-Improved \citep{auer2010ucb}, Median Elimination \citep{even2006action}, Thompson Sampling (TS) \citep{agrawal2011analysis}, OCUCB \citep{lattimore2015optimally}, Bayes-UCB (BU) \citep{kaufmann2012bayesian} and UCB-V \citep{audibert2009exploration}\footnote{The implementation for KLUCB, Bayes-UCB and DMED were taken from \citet{CapGarKau12}}. The parameters of EUCBV algorithm for all the experiments are set as follows: $\psi=\frac{T}{K^2}$ and $\rho =0.5$ (as in Corollary \ref{Result:Corollary:1}). Note that KLUCB+ empirically outperforms KLUCB (see  \citet{garivier2011kl}).

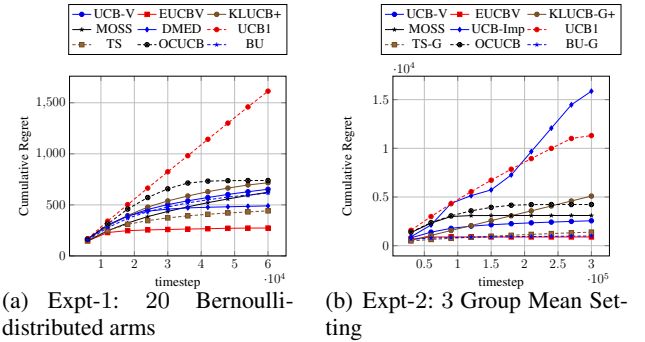
\begin{figure}[!h]
    \centering
    \begin{tabular}{cc}
    \setlength{\tabcolsep}{0.1pt}
    \subfigure[0.25\textwidth][Expt-$1$: $20$ Bernoulli-distributed arms ]
    {
    		\pgfplotsset{
		tick label style={font=\Large},
		label style={font=\Large},
		legend style={font=\Large},
		ylabel style={yshift=5pt},
		}
        \begin{tikzpicture}[scale=0.42]
      	\begin{axis}[
		xlabel={timestep},
		ylabel={Cumulative Regret},
		grid=major,
        clip=true,
  		legend style={at={(0.5,1.4)},anchor=north, legend columns=3} ]
		\addplot table{results/NewExpt/Expt1/UCBV01_comp_subsampled.txt};
		\addplot table{results/NewExpt/Expt1/EUCBV01_comp_subsampled.txt};
		\addplot table{results/NewExpt/Expt1/KLUCB01_comp_subsampled.txt};
		\addplot table{results/NewExpt/Expt1/MOSS01_comp_subsampled.txt};
		\addplot table{results/NewExpt/Expt1/DMED01_comp_subsampled.txt};
		\addplot table{results/NewExpt/Expt1/UCB01_comp_subsampled.txt};
		\addplot table{results/NewExpt/Expt1/TS01_comp_subsampled.txt};
		\addplot table{results/NewExpt/Expt1/OCUCB01_comp_subsampled.txt};
		\addplot table{results/NewExpt/Expt1/BU01_comp_subsampled.txt};
      	\legend{UCB-V,EUCBV,KLUCB+,MOSS,DMED,UCB1,TS,OCUCB,BU}      	
      	\end{axis}
      	\end{tikzpicture}
  		\label{fig:1}
    }
    &
    \subfigure[0.25\textwidth][Expt-$2$: $3$ Group Mean Setting ]
    {
    		\pgfplotsset{
		tick label style={font=\Large},
		label style={font=\Large},
		legend style={font=\Large},
		ylabel style={yshift=5pt},
		}
        \begin{tikzpicture}[scale=0.42]
        \begin{axis}[
		xlabel={timestep},
		ylabel={Cumulative Regret},
       	grid=major,
       	clip=true,
  		legend style={at={(0.5,1.4)},anchor=north, legend columns=3} ]
		\addplot table{results/NewExpt/Expt2/UCBV01_comp_subsampled.txt};
		\addplot table{results/NewExpt/Expt2/EUCBV01_comp_subsampled.txt};
		\addplot table{results/NewExpt/Expt2/KLUCB01_comp_subsampled.txt};
		\addplot table{results/NewExpt/Expt2/MOSS01_comp_subsampled.txt};
		\addplot table{results/NewExpt/Expt2/UCBR01_comp_subsampled.txt};
		\addplot table{results/NewExpt/Expt2/UCB01_comp_subsampled.txt};
		\addplot table{results/NewExpt/Expt2/TS01_comp_subsampled.txt};
		\addplot table{results/NewExpt/Expt2/OCUCB01_comp_subsampled.txt};
		\addplot table{results/NewExpt/Expt2/BU01_comp_subsampled.txt};      	
      	\legend{UCB-V,EUCBV,KLUCB-G+,MOSS,UCB-Imp,UCB1,TS-G,OCUCB,BU-G}
      	\end{axis}
      	\end{tikzpicture}
   		\label{fig:2}
    }
    \end{tabular}
    \caption{A comparison of the cumulative regret incurred by the various bandit algorithms. }
    \label{fig:karmed}
    \vspace*{-1em}
\end{figure}

\textbf{Experiment-1 (Bernoulli with uniform gaps):} This experiment is conducted to observe the performance of EUCBV over a short horizon. The horizon $T$ is set to $60000$. The testbed comprises of $20$ Bernoulli distributed arms with expected rewards of the arms as $r_{1:19}=0.07$ and $r^{*}_{20}=0.1$ and these type of cases are frequently encountered in web-advertising domain (see \cite{garivier2011kl}). The regret is averaged over $100$ independent runs and is shown in Figure \ref{fig:1}. EUCBV, MOSS, OCUCB, UCB1, UCB-V, KLUCB+, TS, BU and DMED are run in this experimental setup. Not only do we observe that EUCBV performs better than all the non-variance based algorithms such as MOSS, OCUCB, UCB-Improved and UCB1, but it also outperforms UCBV because of the choice of the exploration parameters. Because of the small gaps and short horizon $T$, we do not compare with UCB-Improved and Median Elimination. 

\textbf{Experiment-2 (Gaussian $3$ Group Mean Setting):} This experiment is conducted to observe the performance of EUCBV over a large horizon in Gaussian distribution testbed. This setting comprises of a large horizon of $T = 3\times 10^{5}$ timesteps and a large set of arms. This testbed comprises of $100$ arms involving Gaussian reward distributions with expected rewards of the arms in $3$ groups, $r_{1:66}=0.07$, $r_{67:99}=0.01$ and $r^{*}_{100}=0.09$ with variance set as $\sigma_{1:66}^{2} = 0.01,\sigma_{67:99}^{2} = 0.25$ and $\sigma^{2}_{100}=0.25$. The regret is averaged over $100$ independent runs and is shown in Figure \ref{fig:2}. From the results in Figure \ref{fig:2}, we observe that since the gaps are small and the variances of the optimal arm and the arms farthest from the optimal arm are the highest, EUCBV, which allocates pulls proportional to the variances of the arms, outperforms all the non-variance based algorithms MOSS, OCUCB, UCB1, UCB-Improved and Median-Elimination ($\epsilon=0.1,\delta=0.1$). The performance of Median-Elimination is extremely weak in comparison with the other algorithms and its plot is not shown in Figure \ref{fig:2}. We omit its plot in order to more clearly show the difference between EUCBV, MOSS and OCUCB. Also note that the order of magnitude in the y-axis (cumulative regret) of Figure \ref{fig:2} is $10^4$. KLUCB-Gauss+ (denoted by KLUCB-G+), TS-G and BU-G are initialized with Gaussian priors. Both KLUCB-G+ and UCBV which is a variance-aware algorithm perform much worse than TS-G and EUCBV. The performance of DMED is similar to KLUCB-G+ in this setup and its plot is omitted.

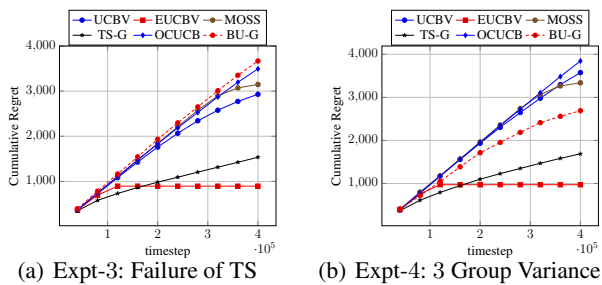
\begin{figure}[!h]
    \centering
    \begin{tabular}{cc}
    \subfigure[0.25\textwidth][Expt-$3$: Failure of TS]
    {
    		\pgfplotsset{
		tick label style={font=\Large},
		label style={font=\Large},
		legend style={font=\Large},
		ylabel style={yshift=5pt},
		}
        \begin{tikzpicture}[scale=0.42]
      	\begin{axis}[
		ylabel={Cumulative Regret},
		xlabel={timestep},
		grid=major,
        clip=true,
  		legend style={at={(0.5,1.2)},anchor=north, legend columns=3} ]
		\addplot table{results/NewExpt/Expt3/UCBV01_comp_subsampled.txt};
		\addplot table{results/NewExpt/Expt3/EUCBV01_comp_subsampled.txt};
		\addplot table{results/NewExpt/Expt3/MOSS01_comp_subsampled.txt};
		\addplot table{results/NewExpt/Expt3/TS01_comp_subsampled.txt};
		\addplot table{results/NewExpt/Expt3/OCUCB01_comp_subsampled.txt};
		\addplot table{results/NewExpt/Expt3/BU01_comp_subsampled.txt};
      	\legend{UCBV,EUCBV,MOSS,TS-G,OCUCB,BU-G} 
      	\end{axis}
      	\end{tikzpicture}
  		\label{fig:3}
    }
    &
    \subfigure[0.25\textwidth][Expt-$4$: $3$ Group Variance]
    {
    	\pgfplotsset{
		tick label style={font=\Large},
		label style={font=\Large},
		legend style={font=\Large},
		ylabel style={yshift=5pt},
		}
        \begin{tikzpicture}[scale=0.42]
        \begin{axis}[
		xlabel={timestep},
		ylabel={Cumulative Regret},
		grid=major,
		clip=true,
  		legend style={at={(0.5,1.2)},anchor=north, legend columns=3} ]
		\addplot table{results/NewExpt/Expt41/UCBV01_comp_subsampled.txt};
		\addplot table{results/NewExpt/Expt41/EUCBV01_comp_subsampled.txt};
		\addplot table{results/NewExpt/Expt41/MOSS01_comp_subsampled.txt};
		\addplot table{results/NewExpt/Expt41/TS01_comp_subsampled.txt};
		\addplot table{results/NewExpt/Expt41/OCUCB01_comp_subsampled.txt};
		\addplot table{results/NewExpt/Expt41/BU01_comp_subsampled.txt};
      	\legend{UCBV,EUCBV,MOSS,TS-G,OCUCB,BU-G} 
      	\end{axis}
        \end{tikzpicture}
        \label{fig:4}
    }
	\end{tabular}
	\label{fig:furtherExpt1}
    \caption{Further Experiments with EUCBV}
    \vspace*{-1em}
\end{figure}


\textbf{Experiment-3 (Failure of TS):} This experiment is conducted to demonstrate that in certain environments when the horizon is large, gaps are small and the variance of the optimal arm is high, the Bayesian algorithms (like TS) do not perform well but EUCBV performs exceptionally well. This experiment is conducted on $100$ Gaussian distributed arms such that expected rewards of the arms $r_{1:10}=0.045$, $r_{11:99}=0.04$, $r^{*}_{100}=0.05$ and the variance is set as $\sigma_{1:10}^{2}=0.01$,   $\sigma_{100}^{2}=0.25$ and $T=4\times 10^5$. The variance of the arms $i=11:99$ are chosen uniform randomly between $[0.2,0.24]$. TS and BU with Gaussian priors fail because here the chosen variance values are such that only variance-aware algorithms with appropriate exploration factors will perform  well or otherwise it will get bogged down in costly exploration. The algorithms that are not variance-aware will spend a significant amount of pulls trying to find the optimal arm. The result is shown in Figure \ref{fig:3}. Predictably EUCBV, which allocates pulls proportional to the variance of the arms, outperforms its closest competitors TS-G, BU-G, UCBV, MOSS and OCUCB. The plots for KLUCB-G+, DMED, UCB1, UCB-Improved and Median Elimination are omitted from the figure as their performance is extremely weak in comparison with other algorithms. We omit their plots to clearly show how EUCBV outperforms its nearest competitors. Note that EUCBV by virtue of its aggressive exploration parameters outperforms UCBV in all the experiments even though UCBV is a variance-based algorithm. The performance of TS-G is also weak and this is in line with the observation in \citet{lattimore2015optimally} that the worst case regret of TS when Gaussian prior is used is $\Omega\left( \sqrt{KT\log T}\right)$.

\textbf{Experiment-4 (Gaussian $3$ Group Variance setting):} This experiment is conducted to show that when the gaps are uniform and variance of the arms is the only discriminative factor then the EUCBV performs extremely well over a very large horizon and over a large number of arms. This testbed comprises of $100$ arms with Gaussian reward distributions, where the expected rewards of the arms are $r_{1:99}=0.09$ and $r^{*}_{100}=0.1$. The variances of the arms are divided into $3$ groups. The group $1$ consist of arms $i=1:49$ where the variances are chosen uniform randomly between $[0.0,0.05]$, group $2$ consist of arms $i=50:99$ where the variances are chosen uniform randomly between $[0.19,0.24]$ and for the optimal arm $i=100$ (group $3$) the variance is set as $\sigma_{*}^{2}=0.25$. We report the cumulative regret averaged over $100$ independent runs. The horizon is set at $T=4\times 10^{5}$ timesteps. We report the performance of MOSS,BU-G, UCBV, TS-G and OCUCB who are the closest competitors of EUCBV over this uniform gap setup. From the results in Figure \ref{fig:4}, it is evident that the growth of regret for EUCBV  is much lower than that of TS-G, MOSS, BU-G, OCUCB and UCBV. Because of the poor performance of KLUCB-G+ in the last two experiments we do not implement it in this setup. Also, note that for optimal performance BU-G, TS-G and KLUCB-G+ require the knowledge of the type of distribution to set their priors . Also, in all the experiments with Gaussian distributions EUCBV significantly outperforms all the Bayesian algorithms initialized with Gaussian priors.

\vspace*{-2mm}
\section{Conclusion and Future Works}
\label{sec:conc}
In this paper, we studied the EUCBV algorithm which takes into account the empirical variance of the arms and employs aggressive exploration parameters in conjunction with non-uniform arm selection (as opposed to UCB-Improved) to eliminate sub-optimal arms. Our theoretical analysis conclusively established that EUCBV exhibits an order-optimal gap-independent regret bound of $O(\sqrt{KT})$. Empirically, we show that EUCBV performs superbly across diverse experimental settings and outperforms most of the bandit algorithms in a stochastic MAB setup. Our experiments show that EUCBV is extremely stable for large horizons and performs consistently well across different types of distributions. One avenue for future work is to remove the constraint of $T\geq K^{2.4}$ required for EUCBV to reach the order optimal regret bound. Another future direction is to come up with an anytime version of EUCBV which does not require horizon $T$ as input parameter. \\

\textbf{Acknowledgements:} This work is supported by a funding from Robert
Bosch Centre for Data Science and Artificial Intelligence (RBC-DSAI) at IIT Madras. The work of the second author is supported by an INSPIRE Faculty Award of the Department of Science and Technology, Govt. of India. 

%


\bibliography{biblio}
\bibliographystyle{aaai}

\clearpage
\newpage

\section{Appendix}
\label{sec:app}

\subsection{Proof of Lemma \ref{proofTheorem:Lemma:1}} 

\label{App:Lemma:1}

\begin{customlem}{1}
If $T\geq K^{2.4}$, $\psi=\dfrac{T}{ K^2}$, $\rho=\dfrac{1}{2}$ and $m\leq \dfrac{1}{2} \log_2\left(\dfrac{T}{e}\right) $, then,
\begin{align*}
\dfrac{\rho m \log(2)}{\log(\psi T) - 2m\log( 2)} \leq \frac{3}{2}.
\end{align*}
\end{customlem}

\begin{proof}
The proof is based on contradiction. Suppose
\begin{eqnarray*}
\dfrac{\rho m \log(2)}{\log(\psi T) - 2m\log( 2)} > \frac{3}{2}.
\end{eqnarray*}
Then, with $\psi=\dfrac{T}{ K^2}$ and $\rho=\dfrac{1}{2}$, we obtain
\begin{eqnarray*}
6\log(K) 
&>& 6\log(T) - 7m\log(2) \\
&\overset{(a)}{\ge}& 6\log(T) - \frac{7}{2} \log_2\left(\frac{T}{e}\right) \log(2) \\
&=& 2.5\log(T) + 3.5 \log_2(e)\log(2)  \\
&\overset{(b)}{=}& 2.5\log(T) +3.5
\end{eqnarray*}
where $(a)$ is obtained using $m\leq \dfrac{1}{2} \log_2\left(\dfrac{T}{e}\right)$, while $(b)$ follows from the identity $\log_2(e)\log(2) =1$. Finally, for $T\ge K^{2.4}$ we obtain, $6\log(K)>6\log(K)+3.5$, which is a contradiction.
\hfill $\blacksquare$	
\end{proof}

\subsection{Proof of Lemma \ref{proofTheorem:Lemma:2}}
\label{App:Lemma:2}
\begin{customlem}{2}
If $T\geq K^{2.4}$, $\psi=\dfrac{T}{ K^2}$, $\rho =\dfrac{1}{2}$, $m_i = min\lbrace m|\sqrt{4\epsilon_{m} } < \dfrac{\Delta_i}{4} \rbrace $ and $c_{i} =\sqrt{\frac{\rho (\hat{v}_i + 2)\log (\psi T\epsilon_{m_{i}})}{4 z_i}}$, then, 
\begin{align*}
c_{i} < \dfrac{\Delta_i}{4}
\end{align*}

\end{customlem}

\begin{proof}

	In the $m_i$-th round since $z_i\geq n_{m_i}$, by substituting $z_i$ with $n_{m_i}$ we can show that, 

\begin{align*}
	c_{i} &\leq \sqrt{\dfrac{\rho (\hat{v}_i + 2)\epsilon_{m_{i}}\log (\psi T\epsilon_{m_{i}})}{2\log(\psi T\epsilon_{m_{i}}^{2})}} \overset{(a)}{\leq} \sqrt{\dfrac{2\rho\epsilon_{m_{i}}\log (\frac{\psi T\epsilon_{m_{i}}^{2}}{\epsilon_{m_{i}}})}{\log(\psi T\epsilon_{m_{i}}^{2})}} \\
	& = \sqrt{\dfrac{2\rho\epsilon_{m_{i}}\log (\psi T\epsilon_{m_{i}}^{2}) - 2\rho\epsilon_{m_{i}}\log (\epsilon_{m_{i}})}{\log(\psi T\epsilon_{m_{i}}^{2})}} \\
	& \leq  \sqrt{2\rho\epsilon_{m_{i}} - \dfrac{2\rho\epsilon_{m_i}\log(\frac{1}{2^{m_i}})}{\log(\psi T \frac{1}{2^{2m_i}})}} \\
	&\leq \sqrt{2\rho\epsilon_{m_{i}} + \dfrac{2\rho\epsilon_{m_i}\log(2^{m_i})}{\log(\psi T) - \log( 2^{2m_i})}}\\
	& \leq \sqrt{2\rho\epsilon_{m_{i}} + \dfrac{2\rho\epsilon_{m_i}m_i \log(2)}{\log(\psi T) - 2m_i\log( 2)}} \\ 
	 & \overset{(b)}{\leq} \sqrt{2\rho\epsilon_{m_{i}} + 2.\frac{3}{2}\epsilon_{m_i}} 
	  < \sqrt{4\epsilon_{m_i}} < \dfrac{\Delta_{i}}{4}.
	\end{align*}
In the above simplification, $(a)$ is due to $\hat{v}_i \in [0,1]$, while $(b)$ is obtained using Lemma~\ref{proofTheorem:Lemma:1}.
\hfill $\blacksquare$	
\end{proof}

\subsection{Proof of Lemma \ref{proofTheorem:Lemma:3}}
\label{App:Lemma:3}

\begin{customlem}{3}
If $m_i = min\lbrace m|\sqrt{4\epsilon_{m} } < \frac{\Delta_i}{4} \rbrace $,  $c_{i} = \sqrt{\frac{\rho (\hat{v}_i + 2) \log (\psi T\epsilon_{m_{i}})}{4 z_{i}}}$ and $n_{m_i} = \frac{\log{(\psi T\epsilon_{m_{i}})}}{2\epsilon_{m_{i}}}$ then we can show that in the $m_i$-th round,
\begin{align*}
\mathbb{P}(\hat{r}_{i}> r_{i} + c_{i})\le \dfrac{2}{(\psi  T\epsilon_{m_{i}})^{\frac{3\rho}{2}}}.
\end{align*}
\end{customlem}


\begin{proof}

We start by recalling from equation (\ref{eq:prob_eq2}) that,

\begin{align}
\mathbb{P}(\hat{r}_{i}> r_{i} + c_{i})
&\leq \mathbb{P}\left( \hat{r}_{i} > r_{i}+ \bar{c}_i\right) 
+ \mathbb{P}\left( \hat{v}_{i}\geq \sigma_{i}^{2}+\sqrt{\epsilon_{m_{i}}}\right)\label{eq:prob_eq3}
\end{align}
where 
\begin{align*}
&c_i =\sqrt{\frac{\rho (\hat{v}_i + 2)\log (\psi T\epsilon_{m_{i}})}{4 z_i}} \text{ and } \\
&\bar{c}_i=\sqrt{\dfrac{\rho (\sigma_{i}^{2}+\sqrt{\epsilon_{m_{i}}} + 2)\log(\psi T\epsilon_{m_{i}})}{4 z_i}}.
\end{align*}

Note that, substituting $ z_i \geq n_{m_i} \geq \frac{\log{(\psi T\epsilon_{m_{i}})}}{2\epsilon_{m_{i}}}$, $\bar{c}_i$ can be simplified to obtain,
\begin{align}
\bar{c}_i
\leq \sqrt{\dfrac{\rho\epsilon_{m_{i}}(\sigma_{i}^{2}+\sqrt{\epsilon_{m_{i}}} + 2)}{2}}\leq \sqrt{ \epsilon_{m_{i}}}.
\label{si_bar_equn}
\end{align}
The first term in the LHS of (\ref{eq:prob_eq3}) can be bounded using the Bernstein inequality as below:
\begin{align}
&\mathbb{P}\left( \hat{r}_{i} > r_{i}+ \bar{c}_i\right)\nonumber 
\le \exp\left(- \dfrac{(\bar{c}_i)^2 z_{i}}{2\sigma_i^2 + \frac{2}{3}\bar{c}_i} \right)\nonumber 
\\
& \overset{(a)}{\le} \exp\left(- \rho \left(\dfrac{3\sigma_{i}^{2}+3\sqrt{\epsilon_{m_{i}}} + 6}{6\sigma_i^2 + 2\sqrt{\epsilon_{m_i}}} \right)\log(\psi  T\epsilon_{m_{i}}\right)\nonumber \\
& \overset{(b)}{\leq} \exp\left(- \rho \log(\psi  T\epsilon_{m_{i}})\right) 
\le \dfrac{1}{(\psi  T\epsilon_{m_{i}})^{\frac{3\rho}{2}}}
\label{lhs1_equn}
\end{align}
where, $(a)$ is obtained by substituting equation \ref{si_bar_equn} and $(b)$ occurs because for all $\sigma_{i}^2 \in [0,\frac{1}{4}]$, $\left(\frac{3\sigma_{i}^{2}+3\sqrt{\epsilon_{m_{i}}} + 6}{6\sigma_i^2 + 2\sqrt{\epsilon_{m_i}}}\right) \geq \frac{3}{2}$ .

The second term in the LHS of (\ref{eq:prob_eq3}) can be simplified as follows:
\begin{align}
&\mathbb{P}\bigg\lbrace \hat{v}_{i}\geq \sigma_{i}^{2}+\sqrt{\epsilon_{m_{i}}}\bigg\rbrace\nonumber\\
&\leq \mathbb{P}\bigg\lbrace \dfrac{1}{n_{i}}\sum_{t=1}^{n_{i}}(X_{i,t}-r_{i})^{2}-(\hat{r}_{i}-r_{i})^{2}\geq \sigma_{i}^{2}+\sqrt{\epsilon_{m_{i}}}\bigg\rbrace\nonumber\\
&\leq \mathbb{P}\bigg\lbrace \dfrac{\sum_{t=1}^{n_{i}}(X_{i,t}-r_{i})^{2}}{n_{i}}\geq \sigma_{i}^{2}+\sqrt{\epsilon_{m_{i}}} \bigg\rbrace\nonumber\\
&\overset{(a)}{\leq} \mathbb{P}\bigg\lbrace \dfrac{\sum_{t=1}^{n_{i}}(X_{i,t}-r_{i})^{2}}{n_{i}}\geq \sigma_{i}^{2} + \bar{c}_i\bigg\rbrace \nonumber\\
&\overset{(b)}{\leq} \exp\left(- \rho \left(\dfrac{3\sigma_{i}^{2}+3\sqrt{\epsilon_{m_{i}}} + 6}{6\sigma_i^2 + 2\sqrt{\epsilon_{m_i}}} \right)\log(\psi  T\epsilon_{m_{i}})\right)
\le \dfrac{1}{(\psi  T\epsilon_{m_{i}})^{\frac{3\rho}{2}}}
\label{lhs2_equn}
\end{align}
where inequality $(a)$ is obtained using (\ref{si_bar_equn}), while $(b)$ follows from the Bernstein inequality. 

Thus, using (\ref{lhs1_equn}) and (\ref{lhs2_equn}) in (\ref{eq:prob_eq3}) we obtain $\mathbb{P}(\hat{r}_{i}> r_{i} + c_{i})\le \dfrac{2}{(\psi  T\epsilon_{m_{i}})^{\frac{3\rho}{2}}}$.
\hfill $\blacksquare$	
\end{proof}

\subsection{Proof of Lemma \ref{proofTheorem:Lemma:4}}
\label{App:Lemma:4}
\begin{customlem}{4}
If $m_i = min\lbrace m|\sqrt{4\epsilon_{m} } < \dfrac{\Delta_i}{4} \rbrace $, $\psi=\frac{T}{ K^2}$, $\rho=\frac{1}{2}$,  $c_{i} =\sqrt{\dfrac{\rho(\hat{v}_i + 2)\log (\psi T\epsilon_{m_{i}})}{4 z_{i}}}$ and $n_{m_i}=\dfrac{\log{(\psi T\epsilon_{m_{i}}^{2})}}{2\epsilon_{m_{i}}}$ then in the $m_i$-th round, 
\begin{align*}
\Pb\lbrace c^{*} > c_i \rbrace  \leq \dfrac{182 K^4}{T^{\frac{5}{4}}\sqrt{\epsilon_{m_i}}}.
\end{align*}
\end{customlem}

\begin{proof}
From the definition of $c_i$ we know that $c_i\propto \frac{1}{z_i}$ as $\psi$ and $T$ are constants. Therefore in the $m_i$-th round,
\begin{align*}
&\Pb\lbrace c^{*} > c_i \rbrace
\leq  \Pb\lbrace  z^* < z_i  \rbrace \\
&\leq \sum_{m=0}^{m_i}\sum_{z^* =1}^{n_{m}}\sum_{z_i =1}^{n_{m}}\bigg(\Pb\lbrace \hat{r}^* < r^* - c^{*}\rbrace + \Pb\lbrace \hat{r}_i > r_i + c_i\rbrace\bigg)
\end{align*}

%

Now, applying Bernstein inequality and following the same way as in Lemma \ref{proofTheorem:Lemma:3} we can show that,
\begin{align*}
&\Pb\lbrace \hat{r}^* < r^* - c^{*}\rbrace \leq \exp(- \frac{(c^{*})^2}{2\sigma_*^2 + \frac{2 c^{*}}{3}} z^*)\leq \frac{4}{(\psi T\epsilon_{m_i})^{\frac{3\rho}{2}}} \\ 
&\Pb\lbrace \hat{r}_i > r_i + c_i\rbrace \leq \exp(- \frac{(c_{i})^2}{2\sigma_i^2 + \frac{2 c_{i}}{3}} z_i)\leq \frac{4}{(\psi T\epsilon_{m_i})^{\frac{3\rho}{2}}}
\end{align*}

Hence, summing everything up, 
\begin{align*}
&\Pb\lbrace c^{*} > c_i \rbrace \\
&\leq \sum_{m=0}^{m_i}\sum_{z^* =1}^{n_{m}}\sum_{z_i =1}^{n_{m}}\bigg(\Pb\lbrace \hat{r}^* < r^* - c^{*}\rbrace + \Pb\lbrace \hat{r}_i > r_i + c_i\rbrace\bigg)\\
&\overset{(a)}{\leq} \sum_{m=0}^{m_i}|B_{m}|n_{m}\bigg(\Pb\lbrace \hat{r}^* < r^* - c^{*}\rbrace + \Pb\lbrace \hat{r}_i > r_i + c_i\rbrace\bigg)\\
&\overset{(b)}{\leq} \sum_{m=0}^{m_i}\dfrac{4K}{(\psi T \epsilon_{m_i})^{\frac{3\rho}{2}}}\dfrac{\log{(\psi T\epsilon_{m}^{2})}}{2\epsilon_{m}}\times 
\\
&\bigg(\Pb\lbrace \hat{r}^* < r^* - c^{*}\rbrace + \Pb\lbrace \hat{r}_i > r_i + c_i\rbrace\bigg)\\
&\overset{(c)}{\leq} \sum_{m=0}^{m_i}\dfrac{4K}{(\psi T \epsilon_{m})^{\frac{3\rho}{2}}}\dfrac{\log(T)}{\epsilon_{m}}\bigg[\frac{4}{(\psi T\epsilon_{m})^{\frac{3\rho}{2}}} + \frac{4}{(\psi T\epsilon_{m})^{\frac{3\rho}{2}}}  \bigg]\\
&\leq \sum_{m=0}^{m_i}\dfrac{32K\log T}{(\psi T\epsilon_{m})^{3\rho}\epsilon_{m}} \leq 
\dfrac{32K\log T}{(\psi T)^{3\rho}}\sum_{m=0}^{m_i}\dfrac{1}{\epsilon_{m}^{3\rho + 1}} \\
&\overset{(d)}{\leq} \sum_{m=0}^{m_i}\dfrac{32K\log T}{(\psi T)^{3\rho}}\left(\sum_{m=0}^{m_i}\dfrac{1}{\epsilon_{m}}\right)^{3\rho + 1}\\
&\overset{(e)}{\leq} 
\dfrac{32 K\log T}{(\frac{T^2}{K^2})^{\frac{3}{2}}}\bigg[\left( 1 + \dfrac{2(2^{ \frac{1}{2}\log_{2} \frac{T}{e}}-1)}{2-1} \right)^{\frac{5}{2}}\bigg] \\
&\leq \dfrac{182 K^4 T^{\frac{5}{4}}\log T}{T^3} \overset{(f)}{\leq} \dfrac{182 K^4}{T^{\frac{5}{4}}} \overset{(g)}{\leq} \dfrac{182 K^4}{T^{\frac{5}{4}}\sqrt{\epsilon_{m_i}}}
\end{align*}

where, $(a)$ comes from the total pulls allocated for all $i\in B_m$ till the $m$-th round, in $(b)$ the arm count $|B_m|$ can be bounded by using equation $(\ref{eq:arm:elim:c1})$ and then we substitute the value of $n_{m}$, $(c)$ happens by substituting the value of $\psi$ and considering $\epsilon_{m}\in [\sqrt{\frac{e}{T}},1]$, $(d)$ follows as $\frac{1}{\epsilon_{m}}\geq 1,\forall m $, in $(e)$ we use the standard geometric progression formula and then we substitute the values of $\rho$ and $\psi$, $(f)$ follows from the inequality $\log T \leq \sqrt{T}$ and $(g)$ is valid for any $\epsilon_{m_i}\in[\sqrt{\frac{e}{T}},1]$. 

\hfill $\blacksquare$	
\end{proof}

\subsection{Proof of Lemma \ref{proofTheorem:Lemma:5}}
\label{App:Lemma:5}
\begin{customlem}{5}
If $m_i = min\lbrace m|\sqrt{4\epsilon_{m} } < \dfrac{\Delta_i}{4} \rbrace $, $\psi=\frac{T}{ K^2}$, $\rho=\frac{1}{2}$, $c_{i} =\sqrt{\frac{\rho (\hat{v}_i + 2)\log (\psi T\epsilon_{m_{i}})}{4 z_i}}$ and $n_{m_i}=\dfrac{\log{(\psi T\epsilon_{m_{i}}^{2})}}{2\epsilon_{m_{i}}}$ then in the $m_i$-th round, 
\begin{align*}
\Pb\lbrace z_i < n_{m_i} \rbrace \leq \dfrac{182 K^4}{T^{\frac{5}{4}}\sqrt{\epsilon_{m_i}}}.
\end{align*}
\end{customlem}

\begin{proof}
Following a similar argument as in Lemma \ref{proofTheorem:Lemma:4}, we can show that in the $m_i$-th round,
\begin{align*}
&\Pb\lbrace z_i < n_{m_i} \rbrace \\
&\leq \sum_{m=0}^{m_i}\sum_{z_i =1}^{n_{m}}\sum_{z^* =1}^{n_{m}}\bigg(\Pb\lbrace \hat{r}^* > r^* - c^{*}\rbrace + \Pb\lbrace \hat{r}_i < r_i + c_i\rbrace\bigg) \\
&\leq \dfrac{32K\log T}{(\psi T)^{3\rho}}\sum_{m=0}^{m_i}\dfrac{1}{\epsilon_{m}^{3\rho + 1}}\leq \dfrac{182 K^4}{T^{\frac{5}{4}}\sqrt{\epsilon_{m_i}}}.
\end{align*}
\hfill $\blacksquare$	
\end{proof}

\subsection{Proof of Lemma \ref{proofTheorem:Lemma:6}}
\label{App:Lemma:6}
\begin{customlem}{6}
For two integer constants $c_1$ and $c_2$, if $20 c_1 \leq c_2$ then,
\begin{align*}
c_1 \dfrac{4\sigma_i^2 + 4}{\Delta_i}\log\bigg( \dfrac{T\Delta_i^2}{K}\bigg) \leq c_2 \dfrac{\sigma_i^2}{\Delta_i}\log\bigg( \dfrac{T\Delta_i^2}{K}\bigg).
\end{align*}
 
\end{customlem}

\begin{proof}
We again prove this by contradiction. Suppose, 
\begin{align*}
c_1 \dfrac{4\sigma_i^2 + 4}{\Delta_i}\log\bigg( \dfrac{T\Delta_i^2}{K}\bigg) > c_2 \dfrac{\sigma_i^2}{\Delta_i}\log\bigg( \dfrac{T\Delta_i^2}{K}\bigg).
\end{align*}

Further reducing the above two terms we can show that, 

\begin{align*}
& 4c_1\sigma_i^2 + 4c_1 > c_2\sigma_i^2\\
& \Rightarrow 4c_1.\dfrac{1}{4} + 4c_1 \overset{(a)}{>} \dfrac{c_2}{4}\\
& \Rightarrow 20 c_1 > c_2.
\end{align*}

Here, $(a)$ occurs because $0\leq\sigma_i^2 \leq \frac{1}{4},\forall i\in \A$. But, we already know that $20 c_1 \leq c_2$. Hence, 
\begin{align*}
c_1 \dfrac{4\sigma_i^2 + 4}{\Delta_i}\log\bigg( \dfrac{T\Delta_i^2}{K}\bigg) \leq c_2 \dfrac{\sigma_i^2}{\Delta_i}\log\bigg( \dfrac{T\Delta_i^2}{K}\bigg).
\end{align*}

\hfill $\blacksquare$	
\end{proof}

\subsection{Proof of Corollary \ref{Result:Corollary:1}}
\label{App:Corollary:1}

\begin{customCorollary}{1}(\textbf{\textit{Gap-Independent Bound}})
When the gaps of all the sub-optimal arms are identical, i.e., $\Delta_i =\Delta = \sqrt{\frac{K\log K}{T}}>\sqrt{\frac{e}{T}}, \forall i\in \A$ and $C_3$ being an integer constant, the
regret of EUCBV is upper bounded by the following gap-independent expression:
\begin{align*}
	\E[R_{T}]\leq  \dfrac{C_3 K^5}{T^{\frac{1}{4}}} + 80\sqrt{KT}.
\end{align*}	
\end{customCorollary}

\begin{proof}
\label{Proof:Corollary:1}
From \cite{bubeck2011pure}  we know that the function $x\in [0,1]\mapsto x\exp(-Cx^2)$ is  decreasing on $\left[\frac{1}{\sqrt{2C}},1\right ]$ for any $C>0$. Thus, we take $C=\left\lfloor \frac{T}{e}\right\rfloor$ and choose  $\Delta_{i}=\Delta=\sqrt{\frac{K\log K}{T}}>\sqrt{\frac{e}{T}}$ for all $i$.

First, let us recall the result in Theorem \ref{Result:Theorem:1} below:
\begin{align*}
\E [R_{T}] \leq &\sum\limits_{i\in \A :\Delta_{i} > b}\bigg\lbrace \dfrac{C_0 K^{4}}{T^{\frac{1}{4}}} + \bigg(\Delta_{i}+\dfrac{320\sigma_i^2\log{(\frac{T\Delta_{i}^{2}}{K})}}{\Delta_{i}}\bigg)\bigg \rbrace\\ 
  & +\sum\limits_{i\in \A :0 < \Delta_{i}\leq b} \dfrac{C_2 K^{4}}{T^{\frac{1}{4}}} + \max_{i\in \A :0 < \Delta_{i}\leq b}\Delta_{i}T.
\end{align*}

Now,  with  $\Delta_i =\Delta = \sqrt{\frac{K\log K}{T}}>\sqrt{\frac{e}{T}}$ we obtain,
	\begin{align*}
	&\sum_{i\in \A :\Delta_{i} > b}\dfrac{320\sigma_i^2\log{(\frac{T\Delta_{i}^{2}}{K})}}{\Delta_{i}} \leq  \dfrac{320\sigma_{\max}^2 K\sqrt{T}\log{(T\dfrac{K(\log K)}{T K})}}{\sqrt{K\log K}}\\ 
	&\leq  \dfrac{320\sigma_{\max}^2\sqrt{KT}\log{(\log K)}}{\sqrt{\log K}}
	\overset{(a)}{\leq} 320\sigma_{\max}^2\sqrt{KT} 
	\end{align*}		
	where $(a)$ follows from the identity $\dfrac{\log{(\log K)}}{\sqrt{\log K}}\leq 1$ for $K\geq 2$. 
	
%

Thus, the total worst case gap-independent bound is given by
	\begin{align*}
	\E[R_{T}] &\overset{(a)}{\leq}  \dfrac{C_3 K^5}{T^{\frac{1}{4}}} + 320\sigma_{\max}^2\sqrt{KT}\\
	&\overset{(b)}{\leq} \dfrac{C_3 K^5}{T^{\frac{1}{4}}} + 80\sqrt{KT}
	\end{align*}	
	
where, in$(a)$, $C_3$ is an integer constant such that $C_3 = C_0 + C_2 $ and $(b)$ occurs because $\sigma_i^2 \in [0,\frac{1}{4}], \forall i\in \A$.

\hfill $\blacksquare$	
\end{proof}

%
%
%

\end{document}